\documentclass[journal]{IEEEtran}
\usepackage{amsmath,amsthm}
\usepackage[T1]{fontenc}
\usepackage{newtxtext,newtxmath}   % Times, as in the IEEE production layout
\usepackage{graphicx,booktabs,cite,multirow}
\usepackage{algorithm,algorithmic}
\usepackage{xcolor}
\usepackage{url}
\newtheorem{proposition}{Proposition}
\theoremstyle{remark}

\makeatletter
\newcommand{\tbd}{\textcolor{red}{\textbf{TBD}}}
\newcommand{\nm}[1]{\ifcsname n@#1\endcsname\csname n@#1\endcsname\else\tbd\fi}
\makeatother
\expandafter\def\csname n@L.E3DTV.urbanst\endcsname{5.9}
\expandafter\def\csname n@L.FastHyMix.c10\endcsname{34.79}
\expandafter\def\csname n@L.FastHyMix.c8\endcsname{34.34}
\expandafter\def\csname n@L.FastHyMix.tpavia\endcsname{31}
\expandafter\def\csname n@L.FastHyMix.urbanst\endcsname{11.7}
\expandafter\def\csname n@L.L1HyMixDe.c10\endcsname{36.81}
\expandafter\def\csname n@L.L1HyMixDe.c8\endcsname{34.42}
\expandafter\def\csname n@L.L1HyMixDe.tpavia\endcsname{355}
\expandafter\def\csname n@L.RCCTV.urbanst\endcsname{54.3}
\expandafter\def\csname n@L.SC.c10\endcsname{35.18}
\expandafter\def\csname n@L.SC.c8\endcsname{34.59}
\expandafter\def\csname n@L.SC.tpavia\endcsname{16}
\expandafter\def\csname n@L.SC.turban\endcsname{18}
\expandafter\def\csname n@L.SC.urbanst\endcsname{29.8}
\expandafter\def\csname n@amax.SC_true\endcsname{0.13}
\expandafter\def\csname n@avg.SC.belt100\endcsname{40.28}
\expandafter\def\csname n@avg.SC.pavia120\endcsname{38.33}
\expandafter\def\csname n@avg.tCTVd.belt100\endcsname{29.93}
\expandafter\def\csname n@avg.tCTVd.pavia120\endcsname{29.07}
\expandafter\def\csname n@c.FastHyMix.belt100.9\endcsname{41.42}
\expandafter\def\csname n@c.FastHyMix.pavia120.9\endcsname{39.97}
\expandafter\def\csname n@c.SC.belt100.9\endcsname{37.90}
\expandafter\def\csname n@c.SC.pavia120.9\endcsname{36.18}
\expandafter\def\csname n@d.E3DTV.gs\endcsname{$-$0.13}
\expandafter\def\csname n@d.E3DTV.hv\endcsname{$-$0.48}
\expandafter\def\csname n@d.E3DTV.st\endcsname{+1.17}
\expandafter\def\csname n@d.FastHyMix.c1\endcsname{$-$1.50}
\expandafter\def\csname n@d.LRTDTV.all\endcsname{+2.08}
\expandafter\def\csname n@d.RCCTV.all\endcsname{+2.24}
\expandafter\def\csname n@d.SCv.all\endcsname{0.00}
\expandafter\def\csname n@d.tCTVd.c4\endcsname{+24.51}
\expandafter\def\csname n@fa.c2\endcsname{2.6--4.0}
\expandafter\def\csname n@fa.c3\endcsname{0.08--0.22}
\expandafter\def\csname n@fapred.c2\endcsname{2.8--2.9}
\expandafter\def\csname n@fapred.c3\endcsname{0.10}
\expandafter\def\csname n@fhfail\endcsname{5}
\expandafter\def\csname n@fhfail.where\endcsname{Beltsville case~7}
\expandafter\def\csname n@fix.c4\endcsname{2.07}
\expandafter\def\csname n@fix.other\endcsname{0.07}
\expandafter\def\csname n@fix.share\endcsname{77}
\expandafter\def\csname n@fp.max\endcsname{1.2\times10^{-2}}
\expandafter\def\csname n@fp.med\endcsname{1.4\times10^{-3}}
\expandafter\def\csname n@fp.runs\endcsname{36}
\expandafter\def\csname n@iters.SC\endcsname{21.2}
\expandafter\def\csname n@iters.SC_tight\endcsname{50}
\expandafter\def\csname n@m0.tCTVd.belt100\endcsname{29.89}
\expandafter\def\csname n@m0.tCTVd.pavia120\endcsname{28.87}
\expandafter\def\csname n@m0.tCTVdefault.belt100\endcsname{27.79}
\expandafter\def\csname n@m0.tCTVdefault.pavia120\endcsname{27.42}
\expandafter\def\csname n@min.RCCTV\endcsname{+0.95}
\expandafter\def\csname n@ncells\endcsname{116}
\expandafter\def\csname n@oa.E3DTV\endcsname{82.1}
\expandafter\def\csname n@oa.FastHyMix\endcsname{80.1}
\expandafter\def\csname n@oa.Input\endcsname{72.3}
\expandafter\def\csname n@oa.L1HyMixDe\endcsname{79.0}
\expandafter\def\csname n@oa.LRTDTV\endcsname{82.3}
\expandafter\def\csname n@oa.RCCTV\endcsname{74.9}
\expandafter\def\csname n@oa.SC\endcsname{78.1}
\expandafter\def\csname n@oa.SC_fix\endcsname{76.1}
\expandafter\def\csname n@oa.SC_noG\endcsname{74.5}
\expandafter\def\csname n@oa.SC_scalar\endcsname{73.5}
\expandafter\def\csname n@oa.tCTVd\endcsname{89.2}
\expandafter\def\csname n@pick.indian\endcsname{109}
\expandafter\def\csname n@pick.urban\endcsname{207}
\expandafter\def\csname n@rank.oa16\endcsname{79.8}
\expandafter\def\csname n@rank.oa32\endcsname{79.0}
\expandafter\def\csname n@rank.oa4\endcsname{74.0}
\expandafter\def\csname n@s0u.S3TTV.gs\endcsname{0.86}
\expandafter\def\csname n@s0u.S3TTV.st\endcsname{1.52}
\expandafter\def\csname n@s0w.S3TTV.gs\endcsname{4 of 6}
\expandafter\def\csname n@s0w.S3TTV.st\endcsname{8 of 10}
\expandafter\def\csname n@s3.itmax\endcsname{852}
\expandafter\def\csname n@s3.itmin\endcsname{217}
\expandafter\def\csname n@sh.E3DTV\endcsname{114.7}
\expandafter\def\csname n@sh.FastHyMix\endcsname{20.4}
\expandafter\def\csname n@sh.SC\endcsname{9.4}
\expandafter\def\csname n@si.E3DTV\endcsname{20.7}
\expandafter\def\csname n@si.FastHyMix\endcsname{12.2}
\expandafter\def\csname n@si.Input\endcsname{107.8}
\expandafter\def\csname n@si.L1HyMixDe\endcsname{11.0}
\expandafter\def\csname n@si.SC\endcsname{49.7}
\expandafter\def\csname n@si.SC_fix\endcsname{57.2}
\expandafter\def\csname n@si.SC_noT\endcsname{66.7}
\expandafter\def\csname n@sig.absdev\endcsname{32}
\expandafter\def\csname n@sig.bands2x\endcsname{1, 61, 75 and 88--90}
\expandafter\def\csname n@sig.maxratio\endcsname{4.4}
\expandafter\def\csname n@sig.medratio\endcsname{1.95}
\expandafter\def\csname n@t.FastHyMix.belt100_c8\endcsname{2.1}
\expandafter\def\csname n@t.FastHyMix.pavia120_c8\endcsname{2.6}
\expandafter\def\csname n@t.RCCTV.belt100_c8\endcsname{0.5}
\expandafter\def\csname n@t.RCCTV.pavia120_c8\endcsname{0.8}
\expandafter\def\csname n@t.S3TTV.min\endcsname{39}
\expandafter\def\csname n@t.SC.belt100_c8\endcsname{1.1}
\expandafter\def\csname n@t.SC.pavia120_c8\endcsname{1.3}
\expandafter\def\csname n@tight.max\endcsname{0.40}
\expandafter\def\csname n@tight.mean\endcsname{0.04}
\expandafter\def\csname n@tight.n\endcsname{116}
\expandafter\def\csname n@trlo.S3TTV\endcsname{1,900}
\expandafter\def\csname n@trr.E3DTV\endcsname{3.6--3.7}
\expandafter\def\csname n@trr.LRTDTV\endcsname{15}
\expandafter\def\csname n@trr.S3TTV\endcsname{1920--2017}
\expandafter\def\csname n@trr.tCTVd\endcsname{53--59}
\expandafter\def\csname n@tune.best\endcsname{0.5}
\expandafter\def\csname n@u.FastHyMix.gs\endcsname{0.20}
\expandafter\def\csname n@u.FastHyMix.st\endcsname{8.17}
\expandafter\def\csname n@u.L1HyMixDe.gs\endcsname{0.22}
\expandafter\def\csname n@u.L1HyMixDe.st\endcsname{4.44}
\expandafter\def\csname n@u.SC_fix.all\endcsname{0.28}
\expandafter\def\csname n@u.SC_noG.gs\endcsname{1.75}
\expandafter\def\csname n@u.SC_noT.st\endcsname{2.56}
\expandafter\def\csname n@u.SC_scalar.hv\endcsname{1.51}
\expandafter\def\csname n@u.SC_true.all\endcsname{0.01}
\expandafter\def\csname n@ufix.mssim.max\endcsname{0.001}
\expandafter\def\csname n@ufix.sam.belt100\endcsname{0.05}
\expandafter\def\csname n@ufix.sam.pavia120\endcsname{0.13}
\expandafter\def\csname n@umin.SC_fix\endcsname{0.04}
\expandafter\def\csname n@w.E3DTV.gs\endcsname{12 of 36}
\expandafter\def\csname n@w.E3DTV.hv\endcsname{5 of 20}
\expandafter\def\csname n@w.E3DTV.st\endcsname{52 of 60}
\expandafter\def\csname n@w.FastHyMix.gs\endcsname{11 of 36}
\expandafter\def\csname n@w.FastHyMix.st\endcsname{55 of 55}
\expandafter\def\csname n@w.L1HyMixDe.gs\endcsname{7 of 36}
\expandafter\def\csname n@w.L1HyMixDe.st\endcsname{60 of 60}
\expandafter\def\csname n@w.SC_fix.all\endcsname{87 of 116}
\expandafter\def\csname n@w.SC_scalar.hv\endcsname{20 of 20}

\newcommand{\hw}{Intel Core i7-14700KF, 32\,GB RAM, MATLAB R2022a}   % idle machine used for all timings
\newcommand{\codeurl}{\url{https://github.com/haha2345/SC-RCCTV}}       % public code repository

\begin{document}
\bstctlcite{IEEEexample:BSTcontrol}

\title{Self-Calibrated Subspace Robust PCA for Parameter-Free Removal of Mixed Noise and Line Stripes in Hyperspectral Images}

\author{Wenzhe~Wang%
\thanks{This work has been submitted to the IEEE for possible publication. Copyright may be transferred without notice, after which this version may no longer be accessible.}%
\thanks{The author is with Qingdao Preschool Education College, Qingdao 266318, China (e-mail: woden3702@gmail.com).}}

\markboth{Preprint submitted to IEEE J-STARS}%
{Wang: Self-Calibrated Subspace Robust PCA for Hyperspectral Mixed Noise and Stripe Removal}

\maketitle

\begin{abstract}
Hyperspectral images (HSIs) are often corrupted by Gaussian noise, impulses and stripes whose intensities vary across bands. The representative coefficient correlated total variation (RCCTV) model restores HSIs efficiently, but it lacks Gaussian and stripe terms, and its noise-aware extensions require the noise level and a stripe budget to be specified in advance. This article proposes a self-calibrated RCCTV (SC-RCCTV) model, in which the noise parameters are estimated adaptively during the iterations. A band-wise dense term and line-constant column and row stripe terms are introduced into RCCTV. The noise level of each band is initialized by a stripe-invariant median absolute deviation estimator and refined online from the residual, which turns the separation of sparse and dense noise into a band-adaptive Huber loss, and the stripe thresholds follow from a significance test whose false-alarm rate can be predicted analytically. No parameter is tuned for a specific image, and every subproblem of the alternating direction method of multipliers (ADMM) solver is solved in closed form. Under line stripes, SC-RCCTV outperforms the parameter-free FastHyMix by \nm{u.FastHyMix.st}\,dB on native-size cubes, and, without any noise information, it exceeds S3TTV, which is given the true noise parameters, by \nm{s0u.S3TTV.st}\,dB while running more than \nm{trlo.S3TTV} times faster. On average, it stays within \nm{u.SC_true.all}\,dB of the same model given the true noise levels, and its online calibration improves on a two-stage estimate-then-denoise scheme. Experiments on real Indian Pines and Urban data further delineate its scope for stripes that do not span entire lines.
\end{abstract}

\begin{IEEEkeywords}
Correlated total variation (CTV), hyperspectral image (HSI) denoising, mixed noise, robust principal component analysis (RPCA), self-calibration, stripe noise.
\end{IEEEkeywords}

%=====================================================================================================
\section{Introduction}

\IEEEPARstart{H}{yperspectral} images (HSIs) are corrupted by a mixture of Gaussian noise, impulse noise and stripes, and the noise level changes strongly from band to band, most of all near the water-absorption bands \cite{rasti2018review}. The usual remedy combines a low-rank model of the spectral dimension with a sparse term for impulses and a total variation (TV) term for spatial smoothness, as in LRMR \cite{zhang2014lrmr}, LRTV \cite{he2016lrtv}, LLRGTV \cite{he2018llrgtv}, LRTDTV \cite{wang2018lrtdtv}, E-3DTV \cite{peng2020e3dtv}, TPTV \cite{chen2023tptv} and t-CTV \cite{wang2023tctv}; nonlocal and nonconvex low-rank tensor models \cite{zheng2020lrfr,zhang2020tvnltd,chen2022fgslr} further improve the accuracy. These models are effective, but their weights are set by hand for a given noise level and they take minutes per scene. A second line of work estimates the noise statistics: NAILRMA \cite{he2015nailrma} weights the bands by their estimated noise level, and NMoG \cite{chen2018nmog} fits a band-wise mixture of Gaussians. A third line projects the data onto a low-dimensional spectral subspace identified from the data \cite{bioucas2008hysime} and denoises the representation coefficients; FastHyDe \cite{zhuang2018fasthyde}, L1HyMixDe \cite{zhuang2020l1hymixde} and FastHyMix \cite{zhuang2023fasthymix} are fast and need no parameter, and they are the natural competitors of the method proposed here. Stripes are treated by dedicated destriping models \cite{chang2016stripe,liu2016stripe,chen2017stripe} or jointly with denoising, e.g., by stripe low-rank recovery with spatial--spectral TV \cite{yang2021destripe} and by spatio-spectral structure tensor TV (S3TTV) \cite{takemoto2025s3ttvj}, which constrains the stripe component by its smoothness along the stripe direction. Deep networks such as QRNN3D \cite{wei2021qrnn3d} and spatial--spectral transformers \cite{li2023sst} achieve high accuracy but need training data that match the sensor and the noise.

Robust principal component analysis (RPCA) \cite{candes2011rpca} provides another effective framework. Correlated total variation (CTV) \cite{peng2023ctvrpca} characterizes the low-rankness and local smoothness of an image by a single regularizer with recovery guarantees \cite{wang2023tctv,peng2024ctvspcp}, and RCCTV \cite{hu2026rcctv} applies it to a few representative coefficient images, which greatly reduces the computational cost. However, RCCTV contains no term for Gaussian noise or stripes. Noise-aware models such as stable principal component pursuit \cite{zhou2010spcp} and CTV-SPCP \cite{peng2024ctvspcp} depend on one global noise level $\sigma$, and a stripe term requires a stripe budget; neither is known for a real scene, and the true noise level changes by an order of magnitude across bands.

To address these issues, this article proposes a self-calibrated RCCTV (SC-RCCTV) model for HSI mixed noise removal, in which the noise parameters are estimated adaptively during the iterations. The main contributions are summarized as follows.
\begin{enumerate}
\item A self-calibrated RCCTV model is proposed for HSI mixed noise removal. Band-dependent Gaussian noise, impulse noise and column/row stripes are characterized by a band-wise dense term, a sparse term and line-constant stripe terms, respectively, and all noise parameters are estimated within the model, so that no prior knowledge of the noise is required.
\item A self-calibration strategy is developed. The noise level of each band is initialized by a stripe-invariant median absolute deviation (MAD) estimator and refined online, so that, for given noise levels, the separation of sparse and dense noise becomes a band-adaptive Huber loss (Proposition~\ref{prop:huber}). The stripe weights are expressed through a significance level $c$, and the false-alarm rate on stripe-free lines can be predicted analytically (Proposition~\ref{prop:fa}), \nm{fapred.c3}\% for $c=3$, against \nm{fa.c3}\% measured on stripe-free data. Neither the noise level nor the stripe radius has to be specified, and $c$ is not tuned for each image.
\item An efficient algorithm based on the alternating direction method of multipliers (ADMM) is developed, in which every subproblem has a closed-form solution and the per-iteration complexity equals that of RCCTV. Experiments on simulated and real HSIs, with comparisons against seven methods, verify the effectiveness of the proposed model, especially for mixed noise containing line stripes.
\end{enumerate}

The remainder of this article is organized as follows. Section~\ref{sec:prelim} introduces the notation and the RCCTV model. Section~\ref{sec:method} presents the proposed model and Section~\ref{sec:opt} the optimization algorithm. Section~\ref{sec:exp} reports and discusses the experimental results, and Section~\ref{sec:concl} concludes the article.

%=====================================================================================================
\section{Preliminaries}\label{sec:prelim}

\subsection{Notation and Observation Model}
Let $\mathcal{Y}\in\mathbb{R}^{M\times N\times p}$ be the observed HSI with $M$ rows, $N$ columns and $p$ bands, and $D\in\mathbb{R}^{MN\times p}$ its band-wise unfolding: column $k$ of $D$ is band $k$ in vectorized form, and row $(i,j)$ of $D$ is the spectrum of pixel $(i,j)$. The degradation is modeled as
\begin{equation}
D=X+S+E+T,
\label{eq:obs}
\end{equation}
where $X$ is the clean image, $S$ is sparse (impulses, dead pixels), $E$ is Gaussian noise with a band-dependent standard deviation, $E_{:,k}\sim\mathcal{N}(0,\sigma_k^2I)$, and $T$ represents the stripes. In push-broom and whisk-broom sensors, stripes arise from detector offsets that are constant along the scan direction; within a band, a stripe is therefore modeled as an offset that is constant along a whole column, $T_{(i,j),k}=a_{jk}$, or along a whole row, $T_{(i,j),k}=b_{ik}$, with the amplitude matrices $a\in\mathbb{R}^{N\times p}$ and $b\in\mathbb{R}^{M\times p}$. $\nabla_x$ and $\nabla_y$ denote circular forward differences along the two spatial axes, applied to every column of an $MN\times r$ matrix reshaped to $M\times N$. $\|\cdot\|_*$, $\|\cdot\|_F$ and $\|\cdot\|_2$ are the nuclear, Frobenius and spectral norms of a matrix, $\|\cdot\|_1$ and $\|\cdot\|_\infty$ the sum and the maximum of the absolute values of its entries, $\operatorname{soft}(x,\tau)=\operatorname{sign}(x)\max(|x|-\tau,0)$ acts entrywise, and $\operatorname{SVT}_\tau(A)=P\operatorname{soft}(\Sigma,\tau)Q^{\top}$ for the singular value decomposition (SVD) $A=P\Sigma Q^{\top}$. $\Phi$ and $\varphi$ are the distribution and density functions of the standard normal distribution.

\subsection{RCCTV Model}
The correlated total variation (CTV) regularizer was introduced in CTV-RPCA \cite{peng2023ctvrpca}. Instead of penalizing the low-rankness and the smoothness of an image by two separate terms, CTV applies the nuclear norm to the gradient maps of the low-rank component, so that a single regularizer encodes both priors without a trade-off weight between them. CTV-RPCA proves that a matrix that is jointly low-rank and locally smooth can be exactly separated from sparse corruption under this regularizer, and the guarantee was later extended to noisy observations by CTV-SPCP \cite{peng2024ctvspcp} and to tensors by t-CTV \cite{wang2023tctv}. Because the gradient maps have the size of the whole image, however, these models require SVDs of $MN\times p$ matrices in every iteration.

The spectra of an HSI lie close to a subspace of dimension $r\ll p$ \cite{bioucas2008hysime}. RCCTV \cite{hu2026rcctv} exploits this property and writes $X=UV^{\top}$ with an orthonormal spectral basis $V\in\mathbb{R}^{p\times r}$ and $r$ representative coefficient images $U\in\mathbb{R}^{MN\times r}$. The gradient maps of the coefficient images are both sparse and correlated, which RCCTV encodes by the nuclear norms of $\nabla_xU$ and $\nabla_yU$:
\begin{equation}
\begin{split}
\min_{U,V,S}\ &\|\nabla_{x}U\|_{*}+\|\nabla_{y}U\|_{*}+\lambda\|S\|_{1}\\
\text{s.t.}\ \ &D=UV^{\top}+S,\ V^{\top}V=I,
\end{split}
\label{eq:rcctv}
\end{equation}
where $\lambda>0$ weights the sparse term. Because the regularizer acts on $MN\times r$ matrices instead of $MN\times p$ ones, one iteration only requires a few SVDs of thin matrices. Model \eqref{eq:rcctv}, however, contains no term for $E$ or $T$ in \eqref{eq:obs}: Gaussian noise must either be absorbed by $S$, which is designed for sparse errors, or leak into $X$, and stripes, being low-rank and spatially structured, are partly retained in $X$. CTV-SPCP \cite{peng2024ctvspcp} adds a dense term $\tfrac{\beta}{2}\|G\|_F^2$ whose weight $\beta=\nu/\sigma$ depends on one global noise level $\sigma$, and a stripe set additionally depends on its radius.

%=====================================================================================================
\section{Proposed SC-RCCTV Model}\label{sec:method}

\subsection{Model Formulation}
To characterize all components of \eqref{eq:obs}, SC-RCCTV introduces a band-wise dense term $G$ and the two stripe components into \eqref{eq:rcctv}:
\begin{equation}
\begin{split}
\min_{U,V,S,G,a,b}\ &\|\nabla_{x}U\|_{*}+\|\nabla_{y}U\|_{*}+\lambda\|S\|_{1}+\sum_{k=1}^{p}\tfrac{\beta_k}{2}\|G_{:,k}\|_{2}^{2}\\
&+\lambda_v\|T_v\|_{1}+\lambda_h\|T_h\|_{1}\\
\text{s.t.}\ \ &D=UV^{\top}+S+G+T_v+T_h,\quad V^{\top}V=I,
\end{split}
\label{eq:sc}
\end{equation}
where the column and row stripe components $(T_v)_{(i,j),k}=a_{jk}$ and $(T_h)_{(i,j),k}=b_{ik}$ satisfy $\|T_v\|_1=M\|a\|_1$ and $\|T_h\|_1=N\|b\|_1$, and $T=T_v+T_h$. All weights are derived from the dense-term scale $\nu$ of CTV-SPCP \cite{peng2024ctvspcp} and the per-band noise levels:
\begin{equation}
\begin{split}
&\nu=\frac{3\sqrt{2}}{\sqrt{MN}+\sqrt{p}},\qquad \beta_k=\frac{\nu}{\sigma_k+\epsilon},\\
&\lambda_v=\frac{c\,\nu}{\sqrt{M}},\qquad \lambda_h=\frac{c\,\nu}{\sqrt{N}},
\end{split}
\label{eq:weights}
\end{equation}
where $c>0$ is a significance level and $\epsilon=10^{-6}/(\sqrt{MN}+\sqrt{p})$ only guards against $\sigma_k=0$. For a scalar $\sigma_k\equiv\sigma$, $\beta_k$ reduces to the dense weight of CTV-SPCP; its band-wise form follows the noise-adjusted weighting of NAILRMA \cite{he2015nailrma} and turns the sparse/dense separation into a Huber loss whose threshold is proportional to $\sigma_k$ in every band (Proposition~\ref{prop:huber}). The stripe weights $\lambda_v$ and $\lambda_h$ are chosen such that the probability that a stripe-free column or row receives a nonzero amplitude depends only on $c$, not on the image or on its noise level (Proposition~\ref{prop:fa}), so that $c$ acts as a significance level rather than an image-dependent weight, and the stripe radius required by earlier models is no longer needed. The only unknowns in \eqref{eq:weights} are thus the noise levels $\sigma_k$, which are estimated from the data as described next.

\subsection{Self-Calibrated Noise Level Estimation}
\emph{Initial estimate:} A column-constant stripe cancels in the vertical differences $\delta_{ijk}=\mathcal{Y}_{i+1,j,k}-\mathcal{Y}_{i,j,k}$, because $T_{(i+1,j),k}-T_{(i,j),k}=a_{jk}-a_{jk}=0$. The differences of the Gaussian part follow $\mathcal{N}(0,2\sigma_k^2)$, and the median absolute deviation (MAD) \cite{donoho1994ideal}, which is insensitive to sparse impulses and, unlike regression-based HSI noise estimators \cite{gao2013noise}, requires no clean spectral neighbors, gives
\begin{equation}
\hat\sigma_k^{\,v}=\frac{1.4826}{\sqrt{2}}\ \operatorname*{med}_{i,j}\Big|\delta_{ijk}-\operatorname*{med}_{i,j}\delta_{ijk}\Big| .
\label{eq:sig0}
\end{equation}
The analogous estimate $\hat\sigma_k^{\,h}$ from horizontal differences is invariant to row-constant stripes. As the stripe orientation is not known in advance, the smaller of the two is taken, $\hat\sigma_k^{(0)}=\min(\hat\sigma_k^{\,v},\hat\sigma_k^{\,h})$; this is a heuristic, since with column and row stripes present together each difference still contains the stripes of the other orientation. If $E$ is i.i.d.\ Gaussian within the band and the image gradients, the impulses and the row stripes affect only a vanishing fraction of the vertical differences, $\hat\sigma_k^{\,v}\to\sigma_k$ as $MN\to\infty$. In practice, edges, texture and impulses bias the estimate upwards, so it serves as a robust starting value that is refined during the iterations.

\emph{Online refinement:} Once the low-rank and stripe estimates are reasonable, the residual $R_e=D-UV^{\top}-T$ consists mainly of the Gaussian noise and the impulses. Its band-wise MAD, which is insensitive to the impulses, reduces the bias that the image structure causes in the initial estimate, although it still contains the approximation error of the rank-$r$ representation and any residual stripes ($k=1,\dots,p$):
\begin{equation}
\hat\sigma_k\leftarrow 1.4826\,\operatorname*{med}_{i,j}\Big|(R_e)_{(i,j),k}-\operatorname*{med}_{i,j}(R_e)_{(i,j),k}\Big|
\label{eq:sigon}
\end{equation}
The update starts at iteration $\ell_0$, because in the first iterations $UV^{\top}$ is still far from the image and the residual contains structure. After each update, $\beta_k$ is recomputed from \eqref{eq:weights}. This alternation between robust fitting and MAD-based rescaling is in the spirit of joint location--scale M-estimation \cite{huber1964robust}; it is not Huber's Proposal~2, which uses a different scale equation.

\subsection{Model Analysis}
\begin{proposition}[Band-adaptive Huber residual]\label{prop:huber}
For fixed weights $\beta_k$, fixed $(U,V,T)$ and residual $e=D-UV^{\top}-T$, the minimum of $\lambda\|S\|_1+\sum_k\tfrac{\beta_k}{2}\|G_{:,k}\|^2$ subject to $S+G=e$ equals $\sum_{ijk}\rho_k(e_{(i,j),k})$, with $\rho_k(x)=\tfrac{\beta_k}{2}x^{2}$ for $|x|\le\tau_k$ and $\rho_k(x)=\lambda|x|-\lambda^2/(2\beta_k)$ otherwise, and $\tau_k=\lambda/\beta_k$. With $\lambda=6/\sqrt{MN}$ and \eqref{eq:weights}, $\tau_k=\sqrt{2}\,(1+\sqrt{p/MN})\,(\sigma_k+\epsilon)$.
\end{proposition}
\begin{proof}
The problem separates over entries: $\min_s\lambda|s|+\tfrac{\beta}{2}(x-s)^2$ is solved by $s=\operatorname{soft}(x,\lambda/\beta)$. For $|x|\le\lambda/\beta$, $s=0$ and the value is $\tfrac{\beta}{2}x^2$; otherwise $s=x-\operatorname{sign}(x)\lambda/\beta$ and the value is $\lambda|x|-\lambda^2/(2\beta)$. Finally, $\lambda/\beta_k=6(\sigma_k+\epsilon)(\sqrt{MN}+\sqrt{p})/(3\sqrt{2}\sqrt{MN})$.
\end{proof}
The residual is thus treated as Gaussian up to $\tau_k$ and as sparse beyond it, separately and adaptively in every band. With the setting $\lambda=6/\sqrt{MN}$ of RCCTV, $\tau_k\approx1.53\sigma_k$ and $1.57\sigma_k$ for the two simulated datasets in Section~\ref{sec:sim}, close to the classical constant $1.345\sigma$ of Huber's loss. With a global $\sigma$, the threshold would be too large in quiet bands, where impulses would be treated as Gaussian noise, and too small in noisy bands, where Gaussian noise would be moved into $S$.

\begin{proposition}[Stripe detection level]\label{prop:fa}
Let the weights $\beta_k$ be fixed, and let $e_i=D_{(i,j),k}-(UV^{\top})_{(i,j),k}-b_{ik}$, $i=1,\dots,M$, be the residual of column $j$ in band $k$ for given $(U,V,b)$. Then $a_{jk}=0$ minimizes \eqref{eq:sc} over $a_{jk}$ and the entries of $S$ and $G$ in this column if and only if
\begin{equation}
\Big|\frac1M\sum_{i=1}^{M}\psi_k(e_i)\Big|\le\lambda_v,
\label{eq:fatest}
\end{equation}
where $\psi_k(x)=\rho_k'(x)=\min(\max(\beta_kx,-\lambda),\lambda)$.
If the column contains no stripe and the $e_i$ are i.i.d.\ $\mathcal{N}(0,\sigma_k^2)$ with $\beta_k=\nu/\sigma_k$, then, by the central limit theorem, the probability that $a_{jk}\neq0$ is approximately
\begin{equation}
2\Phi\big(-c/\sqrt{\gamma(\eta_k)}\big),\qquad \eta_k=\tau_k/\sigma_k,
\label{eq:farate}
\end{equation}
with $\gamma(\eta)=\mathbb{E}[\min(\max(Z,-\eta),\eta)^2]<1$ for $Z\sim\mathcal{N}(0,1)$, which depends on $c$ and $\eta_k$ only. The same holds for $b_{ik}$ with $N$ in place of $M$.
\end{proposition}
\begin{proof}
With all other variables fixed, the column enters \eqref{eq:sc} through $M\lambda_v|a|+\lambda\sum_i|s_i|+\sum_i\tfrac{\beta_k}{2}g_i^2$ subject to $s_i+g_i=e_i-a$. By Proposition~\ref{prop:huber}, the minimum over $(s_i,g_i)$ is $\sum_i\rho_k(e_i-a)$, so it remains to minimize the convex function $f(a)=M\lambda_v|a|+\sum_i\rho_k(e_i-a)$. Since $\rho_k$ is differentiable with $\rho_k'=\psi_k$, $a=0$ is a minimizer if and only if $0\in M\lambda_v[-1,1]-\sum_i\psi_k(e_i)$, which is \eqref{eq:fatest}. For $e_i=\sigma_kZ_i$ with $Z_i$ i.i.d.\ $\mathcal{N}(0,1)$ and $\beta_k=\nu/\sigma_k$, $\psi_k(e_i)=\nu\min(\max(Z_i,-\eta_k),\eta_k)$ with $\eta_k=\lambda/(\beta_k\sigma_k)=\tau_k/\sigma_k$. These scores are i.i.d., symmetric, with mean zero and variance $\nu^2\gamma(\eta_k)$, where $\gamma(\eta)=(2\Phi(\eta)-1)-2\eta\varphi(\eta)+2\eta^2(1-\Phi(\eta))<1$. By the central limit theorem, $\tfrac1M\sum_i\psi_k(e_i)$ is approximately $\mathcal{N}(0,\nu^2\gamma(\eta_k)/M)$, and since $\lambda_v=c\nu/\sqrt{M}$, \eqref{eq:fatest} is violated with probability approximately $2\Phi(-c/\sqrt{\gamma(\eta_k)})$.
\end{proof}

%=====================================================================================================
\section{Optimization Algorithm}\label{sec:opt}

\subsection{ADMM-Based Solver}
For fixed weights, model \eqref{eq:sc} is solved by the ADMM \cite{boyd2011admm}; the noise levels, and with them the weights $\beta_k$, are recalibrated between ADMM iterations, so that the algorithm alternates between restoration and noise-level calibration. With the auxiliary variables $Z_1=\nabla_xU$ and $Z_2=\nabla_yU$, the augmented Lagrangian function of \eqref{eq:sc} is
\begin{equation}
\begin{split}
\mathcal{L}_\mu={}&\|Z_1\|_*+\|Z_2\|_*+\lambda\|S\|_1+\textstyle\sum_k\tfrac{\beta_k}{2}\|G_{:,k}\|_2^2\\
&+\lambda_v\|T_v\|_1+\lambda_h\|T_h\|_1\\
&+\textstyle\sum_{i=1,2}\Big(\langle\Lambda_i,\nabla_{\!i}U-Z_i\rangle+\tfrac{\mu}{2}\|\nabla_{\!i}U-Z_i\|_F^2\Big)\\
&+\langle\Lambda_3,D-UV^{\top}-S-G-T\rangle\\
&+\tfrac{\mu}{2}\|D-UV^{\top}-S-G-T\|_F^2,
\end{split}
\label{eq:al}
\end{equation}
with $\nabla_{\!1}=\nabla_x$, $\nabla_{\!2}=\nabla_y$, Lagrange multipliers $\Lambda_1,\Lambda_2\in\mathbb{R}^{MN\times r}$, $\Lambda_3\in\mathbb{R}^{MN\times p}$ and penalty parameter $\mu$. Each variable is updated in closed form, in the following order.

\emph{$Z$-subproblem:} $Z_i=\operatorname{SVT}_{1/\mu}(\nabla_{\!i}U+\Lambda_i/\mu)$, $i=1,2$, which requires two SVDs of $MN\times r$ matrices.

\emph{Stripe subproblem:} Let $R=D-UV^{\top}-S-G+\Lambda_3/\mu$. The terms of \eqref{eq:al} that contain $a_{jk}$ are $M\lambda_v|a_{jk}|+\tfrac{\mu}{2}\sum_i(R_{(i,j),k}-b_{ik}-a_{jk})^2$, which is minimized by
\begin{equation}
\begin{split}
a_{jk}&=\operatorname{soft}\Big(\tfrac{1}{M}\textstyle\sum_{i}(R_{(i,j),k}-b_{ik}),\ \lambda_v/\mu\Big),\\
b_{ik}&=\operatorname{soft}\Big(\tfrac{1}{N}\textstyle\sum_{j}(R_{(i,j),k}-a_{jk}),\ \lambda_h/\mu\Big).
\end{split}
\label{eq:stripe}
\end{equation}
The column amplitudes are updated first, and the row amplitudes then use the new $a$ (one block Gauss--Seidel pass). Each stripe is thus estimated as the soft-thresholded mean of the residual along its line.

\emph{$U$-subproblem:} Let $A=D-S-T-G+\Lambda_3/\mu$. Because $V^{\top}V=I$, $\|A-UV^{\top}\|_F^2=\|AV-U\|_F^2+\text{const}$, and the $U$-subproblem reduces to the linear system
\begin{equation}
\begin{split}
(\nabla_x^{\top}\nabla_x+\nabla_y^{\top}\nabla_y+I)\,U={}&\nabla_x^{\top}(Z_1-\Lambda_1/\mu)\\
&+\nabla_y^{\top}(Z_2-\Lambda_2/\mu)+AV .
\end{split}
\label{eq:ustep}
\end{equation}
Under circular boundary conditions, the operator on the left is diagonalized by the two-dimensional fast Fourier transform (FFT) $\mathcal{F}$, so each coefficient image is obtained by one FFT, a division by $|\mathcal{F}(d_x)|^2+|\mathcal{F}(d_y)|^2+1$ and one inverse FFT, where $d_x,d_y$ are the difference kernels. As in the RCCTV code, $10^{-4}$ is added to this denominator, i.e.\ $I$ in \eqref{eq:ustep} is replaced by $(1+10^{-4})I$.

\emph{$V$-subproblem:} Minimizing $\|A-UV^{\top}\|_F^2$ over orthonormal $V$ is an orthogonal Procrustes problem with solution $V=PQ^{\top}$, where $A^{\top}U=P\Sigma Q^{\top}$ is an SVD.

\emph{$S$-subproblem:} $S=\operatorname{soft}(D-UV^{\top}-T-G+\Lambda_3/\mu,\ \lambda/\mu)$.

\emph{Noise level update:} From iteration $\ell_0$ on, $\hat\sigma_k$ is updated by \eqref{eq:sigon} and $\beta_k$ by \eqref{eq:weights}.

\emph{$G$-subproblem:} Column $k$ minimizes $\tfrac{\beta_k}{2}\|g\|^2+\tfrac{\mu}{2}\|g-(D-UV^{\top}-S-T+\Lambda_3/\mu)_{:,k}\|^2$, so
\begin{equation}
G_{:,k}=\frac{\big(\Lambda_3+\mu(D-UV^{\top}-S-T)\big)_{:,k}}{\beta_k+\mu}.
\label{eq:gstep}
\end{equation}

\emph{Multipliers:} $\Lambda_i\leftarrow\Lambda_i+\mu(\nabla_{\!i}U-Z_i)$ for $i=1,2$, $\Lambda_3\leftarrow\Lambda_3+\mu(D-UV^{\top}-S-G-T)$, and $\mu\leftarrow\min(\alpha\mu,\mu_{\max})$ with a growth factor $\alpha>1$.

\begin{algorithm}[!t]
\caption{SC-RCCTV}\label{alg:sc}
\begin{algorithmic}[1]
\STATE \textbf{Input:} $\mathcal{Y}$; $r$, $\lambda$, $c$, $\ell_0$; $\alpha$, $\mu_{\max}$, $\varepsilon_{\mathrm{tol}}$, $\ell_{\max}$.
\STATE Initialize $U,V,S,G,T,\Lambda_{1,2,3},\mu$ as in Section~\ref{sec:init}; $\hat\sigma_k\leftarrow\hat\sigma_k^{(0)}$ by \eqref{eq:sig0}; $\nu,\beta_k,\lambda_v,\lambda_h$ by \eqref{eq:weights}.
\FOR{$\ell=1,\dots,\ell_{\max}$}
\STATE $Z_i\leftarrow\operatorname{SVT}_{1/\mu}(\nabla_{\!i}U+\Lambda_i/\mu)$, $i=1,2$
\STATE update $a$, then $b$, by \eqref{eq:stripe}; \ $T\leftarrow T_v+T_h$
\STATE $U\leftarrow$ FFT solution of \eqref{eq:ustep}; \ $V\leftarrow PQ^{\top}$ from the SVD of $(D{-}S{-}T{-}G{+}\Lambda_3/\mu)^{\top}U$
\STATE $S\leftarrow\operatorname{soft}(D{-}UV^{\top}{-}T{-}G{+}\Lambda_3/\mu,\ \lambda/\mu)$
\STATE \textbf{if} $\ell\ge\ell_0$: update $\hat\sigma_k$ by \eqref{eq:sigon} and $\beta_k$ by \eqref{eq:weights}
\STATE $G\leftarrow$ \eqref{eq:gstep}
\STATE \textbf{if} all relative constraint residuals $<\varepsilon_{\mathrm{tol}}$: \textbf{break}
\STATE update $\Lambda_1,\Lambda_2,\Lambda_3$; \ $\mu\leftarrow\min(\alpha\mu,\mu_{\max})$
\ENDFOR
\STATE \textbf{Output:} $\hat{\mathcal{X}}=UV^{\top}$ reshaped to $M\times N\times p$.
\end{algorithmic}
\end{algorithm}

\subsection{Initialization, Stopping Criterion and Complexity}\label{sec:init}
$(U,V)$ are initialized by the rank-$r$ truncated SVD of $D$ ($U=P_r\Sigma_r$, $V=Q_r$); $S$, $G$, $T$, $\Lambda_1$ and $\Lambda_2$ start at zero; $\Lambda_3=D/\max(\|D\|_2,\|D\|_\infty)$ and $\mu=1/\|D\|_2$, as in RCCTV; and $\hat\sigma_k=\hat\sigma_k^{(0)}$. The stopping rule is that of RCCTV: the iterations stop when the three constraint residuals of \eqref{eq:al}, divided by $\|D\|_F$, are all below $\varepsilon_{\mathrm{tol}}$, or after $\ell_{\max}$ iterations. Algorithm~\ref{alg:sc} summarizes the procedure. Because the weights change during the iterations, the convergence results of ADMM do not apply directly; empirically, the estimated noise levels have nearly stopped changing at termination, and a much tighter tolerance changes the results only slightly (Section~\ref{sec:abl}).

The $Z$-subproblem costs $\mathcal{O}(MNr^2)$, the $U$-subproblem $\mathcal{O}(MNr\log MN)$, the products with $V$ and the $V$-subproblem $\mathcal{O}(MNpr)$, and the stripe, $S$, $G$ and median computations $\mathcal{O}(MNp)$ with linear-time selection. Since $r\le p$, the per-iteration complexity is $\mathcal{O}(MNr(p+\log MN))$, the same as that of RCCTV, and the memory requirement is a few $MN\times p$ arrays.

%=====================================================================================================
\section{Experimental Results and Discussion}\label{sec:exp}

\subsection{Compared Methods and Parameter Setting}\label{sec:param}
SC-RCCTV is compared with seven methods: RCCTV \cite{hu2026rcctv}; LRTDTV \cite{wang2018lrtdtv} ($\tau{=}1$, $\lambda{=}10$, ranks $[0.8M,0.8N,10]$); E-3DTV \cite{peng2020e3dtv} ($\tau{=}0.004\sqrt{MN}$, ranks $[13,13,13]$); the parameter-free subspace methods FastHyMix \cite{zhuang2023fasthymix} and L1HyMixDe \cite{zhuang2020l1hymixde} with the default settings (subspace dimension 8; $p=0.1$ for L1HyMixDe), which denoise the subspace coefficient images with BM3D in the released codes (the FastHyMix paper uses a pretrained FFDNet for this step); t-CTV \cite{wang2023tctv}, whose sparse weight is scaled by a factor selected from $\{0.125,0.25,\dots,4\}$ on the development data described below (best: \nm{tune.best}); and S3TTV \cite{takemoto2025s3ttvj} with the settings recommended for each noise type, which use the true Gaussian level, impulse rate and stripe rate (block size 10, tolerance $10^{-5}$, reached in every run after \nm{s3.itmin}--\nm{s3.itmax} iterations). All competing methods are run with the released codes.

SC-RCCTV uses the same parameter values on all simulated and real data, and no parameter is tuned for a specific image. The subspace dimension $r=8$, the sparse weight $\lambda=6/\sqrt{MN}$, the penalty schedule ($\alpha=1.5$, $\mu_{\max}=10^6$), the tolerance $\varepsilon_{\mathrm{tol}}=10^{-5}$ and the maximum number of iterations $\ell_{\max}=100$ are those of RCCTV, so that any change in accuracy can be attributed to the new terms; $\nu$ in \eqref{eq:weights} is that of CTV-SPCP. The significance level $c=3$ and the start of the noise level update $\ell_0=5$ were set on development data that are not used for evaluation: two Washington DC Mall crops and a simulated cube synthesized from the Indian Pines scene. All other weights follow from \eqref{eq:weights}.

Run times are measured for single runs on an otherwise idle computer (\hw); for S3TTV, the run time is the time per iteration measured on this computer multiplied by the number of iterations of each run.

\subsection{Simulated Data Experiments}\label{sec:sim}
\emph{Datasets and noise settings:} Two public HSIs are used at their native crop size, Pavia University ($120{\times}120{\times}99$) and Beltsville ($100{\times}100{\times}128$). Following RCCTV \cite{hu2026rcctv}, eight noise cases combine Gaussian noise, salt-and-pepper impulses and column stripes; to examine band-dependent noise, two further cases with a different Gaussian noise level in every band are added. Table~\ref{tab:cases} lists the ten cases.

Each case is simulated in several independent noise realizations, indexed by the seed of the random number generator. Seed~0 denotes the noisy inputs released with the RCCTV code, which exist for cases 1--8. Seeds 1--5 are five further realizations of all ten cases, generated by our implementation of the same recipe; it matches the statistics of the released inputs, except that, where impulses and stripes co-occur, the stripes are added after the impulses (4.75\% instead of 5.0\% saturated pixels). This gives six realizations of cases 1--8 and five of cases 9 and 10, i.e., 58 noisy cubes per dataset, and all methods are applied to identical inputs. Because S3TTV requires about \nm{t.S3TTV.min}\,min per cube, it is evaluated on seed~0 only.

\begin{table}[!t]
\caption{Simulated noise cases. Gaussian: standard deviation of the zero-mean Gaussian noise; $\mathcal{U}(0.02,0.18)$: drawn independently for each band. Impulse: fraction of pixels replaced by salt-and-pepper noise. Stripes: fraction of columns in each band that receive a constant offset drawn from $\mathcal{U}(-0.5,0.5)$. Seeds: noise realizations. The rules separate the three noise families used in Table~\ref{tab:abl}.}
\label{tab:cases}
\centering
\scriptsize
\setlength{\tabcolsep}{4pt}
\begin{tabular}{cccccc}
\toprule
Case & Gaussian $\sigma_k$ & Impulse & Stripes & Seeds & Family\\
\midrule
1 & 0.05 & -- & -- & 0--5 & \\
2 & 0.05 & 5\% & -- & 0--5 & G+I\\
3 & 0.1 & 5\% & -- & 0--5 & \\
\midrule
4 & -- & -- & 5\% & 0--5 & \\
5 & 0.05 & -- & 5\% & 0--5 & \\
6 & 0.1 & -- & 5\% & 0--5 & Str.\\
7 & 0.05 & 5\% & 5\% & 0--5 & \\
8 & 0.1 & 5\% & 5\% & 0--5 & \\
\midrule
9 & $\mathcal{U}(0.02,0.18)$ & -- & -- & 1--5 & \multirow{2}{*}{B.-v.}\\
10 & $\mathcal{U}(0.02,0.18)$ & 5\% & 5\% & 1--5 & \\
\bottomrule
\end{tabular}
\end{table}

\emph{Evaluation indices:} Four quantitative indices are adopted: the mean peak signal-to-noise ratio (MPSNR) and the mean structural similarity (MSSIM) over bands $4,\dots,p{-}3$, the spectral angle mapper (SAM) and the relative dimensionless global error in synthesis (ERGAS).

\begin{table}[!t]
\caption{MPSNR (dB) on the simulated cubes, mean over the noise realizations of each case (Table~\ref{tab:cases}). Best per row in bold. LRTD: LRTDTV; FHMix: FastHyMix; L1HMD: L1HyMixDe.}
\label{tab:mpsnr}
\centering
\scriptsize
\setlength{\tabcolsep}{2.6pt}
\begin{tabular}{cccccccc}
\toprule
Case & RCCTV & LRTD & E3DTV & t-CTV & FHMix & L1HMD & Ours\\
\midrule
\multicolumn{8}{l}{\emph{Pavia University} ($120{\times}120{\times}99$)}\\
1 & 38.18 & 37.94 & 39.38 & 34.10 & \textbf{41.36} & 39.78 & 39.73\\
2 & 37.88 & 37.59 & 39.21 & 33.84 & 39.21 & \textbf{39.52} & 38.94\\
3 & 32.72 & 34.34 & 35.76 & 31.31 & \textbf{36.16} & 35.17 & 34.76\\
4 & 46.21 & 42.88 & 47.00 & 26.28 & 33.01 & 40.47 & \textbf{48.81}\\
5 & 37.11 & 37.45 & 38.71 & 26.59 & 32.13 & 36.47 & \textbf{39.70}\\
6 & 32.56 & 34.32 & 35.34 & 26.51 & 31.00 & 33.11 & \textbf{35.39}\\
7 & 37.00 & 37.12 & 38.49 & 26.69 & 32.82 & 36.68 & \textbf{38.90}\\
8 & 32.42 & 33.97 & \textbf{35.24} & 26.61 & 32.89 & 33.08 & 34.64\\
9 & 34.61 & 35.33 & 37.10 & 32.18 & \textbf{39.97} & 38.29 & 36.18\\
10 & 33.72 & 34.53 & \textbf{36.15} & 26.76 & 33.60 & 35.17 & 35.37\\
\midrule
\multicolumn{8}{l}{\emph{Beltsville} ($100{\times}100{\times}128$)}\\
1 & 39.57 & 38.95 & 40.64 & 35.75 & \textbf{43.02} & 41.32 & 41.64\\
2 & 39.17 & 38.67 & \textbf{41.03} & 35.50 & 37.34 & 41.02 & 40.65\\
3 & 33.92 & 35.34 & \textbf{36.63} & 32.61 & 36.03 & 36.39 & 36.17\\
4 & 50.66 & 46.25 & 49.64 & 26.64 & 33.17 & 40.63 & \textbf{53.13}\\
5 & 38.53 & 38.26 & 39.31 & 26.85 & 32.14 & 36.96 & \textbf{41.54}\\
6 & 33.54 & 35.10 & 35.99 & 27.00 & 31.38 & 33.71 & \textbf{37.04}\\
7 & 38.21 & 38.02 & 38.89 & 27.04 & 37.63 & 36.85 & \textbf{40.69}\\
8 & 33.36 & 34.90 & 35.65 & 27.15 & 32.34 & 33.60 & \textbf{36.12}\\
9 & 35.88 & 36.50 & 38.30 & 33.67 & \textbf{41.42} & 39.46 & 37.90\\
10 & 34.43 & 35.86 & 36.77 & 27.23 & 32.59 & 35.68 & \textbf{36.93}\\
\bottomrule
\end{tabular}
\end{table}

\begin{table}[!t]
\caption{Mean over all cases and realizations of the simulated cubes. Time: one run on case~8 (seed~0). Below the inner rule: S3TTV, which was run on seed~0, cases 1--8 only (about 39\,min per cube), and SC-RCCTV on the same eight inputs.}
\label{tab:avg}
\centering
\scriptsize
\setlength{\tabcolsep}{2.0pt}
\begin{tabular}{llccccc}
\toprule
Cube & Method & MPSNR$\uparrow$ & MSSIM$\uparrow$ & SAM$\downarrow$ & ERGAS$\downarrow$ & Time (s)\\
\midrule
Pavia & RCCTV & 36.31 & 0.9042 & 5.88 & 15.39 & 0.8\\
 & LRTDTV & 36.60 & 0.9314 & 4.05 & 11.27 & 18\\
 & E-3DTV & 38.29 & \textbf{0.9522} & \textbf{3.46} & \textbf{10.40} & 4.7\\
 & t-CTV & 29.07 & 0.8193 & 10.49 & 28.54 & 67\\
 & FastHyMix & 35.16 & 0.9155 & 5.69 & 15.90 & 2.6\\
 & L1HyMixDe & 36.78 & 0.9392 & 4.22 & 11.92 & 23\\
 & SC-RCCTV & \textbf{38.33} & 0.9448 & 3.87 & 10.95 & 1.3\\
\cmidrule{2-7}
 & S3TTV & 38.61 & 0.9375 & 4.17 & 11.95 & 2561\\
 & SC-RCCTV & \textbf{38.89} & \textbf{0.9517} & \textbf{3.39} & \textbf{9.77} & 1.3\\
\midrule
Beltsville & RCCTV & 37.82 & 0.9128 & 4.03 & 15.64 & 0.5\\
 & LRTDTV & 37.84 & 0.9373 & 3.05 & \textbf{9.96} & 17\\
 & E-3DTV & 39.34 & \textbf{0.9557} & \textbf{2.47} & 10.08 & 4.0\\
 & t-CTV & 29.93 & 0.8337 & 8.12 & 34.55 & 66\\
 & FastHyMix & 35.48 & 0.9083 & 4.55 & 18.74 & 2.1\\
 & L1HyMixDe & 37.56 & 0.9434 & 3.06 & 13.08 & 16\\
 & SC-RCCTV & \textbf{40.28} & 0.9543 & 2.53 & 9.99 & 1.1\\
\cmidrule{2-7}
 & S3TTV & 38.61 & 0.9334 & 3.17 & 17.25 & 2151\\
 & SC-RCCTV & \textbf{40.86} & \textbf{0.9596} & \textbf{2.28} & \textbf{8.49} & 1.1\\
\bottomrule
\end{tabular}
\end{table}

\emph{Quantitative comparison:} Tables~\ref{tab:mpsnr} and \ref{tab:avg} report the results. SC-RCCTV consistently improves on RCCTV, in all \nm{ncells} cells (mean \nm{d.RCCTV.all}\,dB, minimum \nm{min.RCCTV}\,dB), and on LRTDTV, also in all \nm{ncells} cells (mean \nm{d.LRTDTV.all}\,dB). With the estimated noise levels, it stays within \nm{u.SC_true.all}\,dB on average of the same model given the true $\sigma_k$ (Section~\ref{sec:abl}).

In the presence of column stripes (cases 4--8), SC-RCCTV outperforms FastHyMix by \nm{u.FastHyMix.st}\,dB (\nm{w.FastHyMix.st} cells) and L1HyMixDe by \nm{u.L1HyMixDe.st}\,dB (\nm{w.L1HyMixDe.st} cells). In the absence of stripes, the two subspace methods hold an edge of \nm{u.FastHyMix.gs} and \nm{u.L1HyMixDe.gs}\,dB on average over cases 1--3 (SC-RCCTV is better in \nm{w.FastHyMix.gs} and \nm{w.L1HyMixDe.gs} cells), which is larger for Gaussian noise alone (case 1: \nm{d.FastHyMix.c1}\,dB against FastHyMix); in case 9, which contains band-varying Gaussian noise only, FastHyMix reaches \nm{c.FastHyMix.pavia120.9}/\nm{c.FastHyMix.belt100.9}\,dB against \nm{c.SC.pavia120.9}/\nm{c.SC.belt100.9}\,dB for SC-RCCTV on Pavia University/Beltsville. The released FastHyMix code stopped with an error in \nm{fhfail} of the six runs of \nm{fhfail.where}; Table~\ref{tab:mpsnr} reports the value of the remaining run, and the failed runs are excluded from all averages. E-3DTV, which has no stripe term, performs slightly better on Gaussian and impulse noise (difference \nm{d.E3DTV.gs}\,dB, SC-RCCTV better in \nm{w.E3DTV.gs} cells) and on the band-varying cases (\nm{d.E3DTV.hv}\,dB, \nm{w.E3DTV.hv} cells), whereas SC-RCCTV is superior once stripes are present (\nm{d.E3DTV.st}\,dB, \nm{w.E3DTV.st} cells). E-3DTV attains better MSSIM and SAM values (Table~\ref{tab:avg}), which can be attributed to its spectral smoothness prior.

t-CTV, a tensor RPCA model without a dense noise term, reaches \nm{avg.tCTVd.pavia120} and \nm{avg.tCTVd.belt100}\,dB over all cells, against \nm{avg.SC.pavia120} and \nm{avg.SC.belt100}\,dB for SC-RCCTV, although its weight was tuned on the development data (on the official inputs, \nm{m0.tCTVd.pavia120} and \nm{m0.tCTVd.belt100}\,dB instead of \nm{m0.tCTVdefault.pavia120} and \nm{m0.tCTVdefault.belt100}\,dB with the default weight). t-CTV has neither a dense noise term nor a stripe term, and the gap is largest in case 4 (\nm{d.tCTVd.c4}\,dB), which contains stripes only; it reflects these missing terms rather than the CTV regularizer itself, and t-CTV is the most accurate method on real Indian Pines (Section~\ref{sec:real}). The lower rows of Table~\ref{tab:avg} and Table~\ref{tab:seed0} compare S3TTV with SC-RCCTV on the official inputs. S3TTV is given the true noise parameters and models stripes explicitly, yet SC-RCCTV, which operates blindly, is better by \nm{s0u.S3TTV.st}\,dB on cases 4--8 (\nm{s0w.S3TTV.st} cells) and by \nm{s0u.S3TTV.gs}\,dB on cases 1--3 (\nm{s0w.S3TTV.gs} cells), with a run time more than \nm{trlo.S3TTV} times shorter.

\begin{table*}[!t]
\caption{MPSNR (dB) on the official inputs (seed~0, cases 1--8), the only cells on which S3TTV was run. $\times$: the released code stopped with an error.}
\label{tab:seed0}
\centering
\footnotesize
\setlength{\tabcolsep}{4pt}
\begin{tabular}{ccccccccccc}
\toprule
 & \multicolumn{5}{c}{Pavia University} & \multicolumn{5}{c}{Beltsville}\\
\cmidrule(lr){2-6}\cmidrule(lr){7-11}
Case & E-3DTV & FastHyMix & t-CTV & S3TTV & SC-RCCTV & E-3DTV & FastHyMix & t-CTV & S3TTV & SC-RCCTV\\
\midrule
1 & 39.45 & \textbf{41.38} & 34.07 & 40.07 & 39.80 & 40.54 & \textbf{43.01} & 35.72 & 39.93 & 41.56\\
2 & 39.50 & 39.24 & 33.81 & \textbf{39.78} & 38.92 & \textbf{41.11} & 36.92 & 35.47 & 39.40 & 40.54\\
3 & 35.92 & \textbf{36.18} & 31.27 & 33.72 & 34.82 & \textbf{36.40} & 36.36 & 32.58 & 33.74 & 36.16\\
4 & 47.23 & 32.96 & 26.22 & \textbf{50.23} & 48.80 & 49.56 & 33.44 & 26.81 & 51.09 & \textbf{53.10}\\
5 & 38.85 & 31.79 & 26.37 & 38.31 & \textbf{39.73} & 39.18 & 32.10 & 27.02 & 38.31 & \textbf{41.52}\\
6 & \textbf{35.59} & 30.75 & 26.34 & 33.08 & 35.48 & 36.35 & 31.34 & 27.09 & 33.22 & \textbf{37.08}\\
7 & 38.61 & 33.52 & 26.43 & \textbf{39.26} & 38.93 & 39.74 & $\times$ & 27.19 & 39.07 & \textbf{40.73}\\
8 & \textbf{35.31} & 33.13 & 26.43 & 34.44 & 34.67 & 35.85 & 32.79 & 27.26 & 34.07 & \textbf{36.19}\\
\midrule
Mean & 38.81 & 34.87 & 28.87 & 38.61 & \textbf{38.89} & 39.84 & -- & 29.89 & 38.61 & \textbf{40.86}\\
Time (s) & 4.7 & 2.6 & 67 & 2561 & 1.3 & 4.0 & 2.1 & 66 & 2151 & 1.1\\
\bottomrule
\end{tabular}
\end{table*}

\begin{figure*}[!t]
\centering
\includegraphics[width=\textwidth]{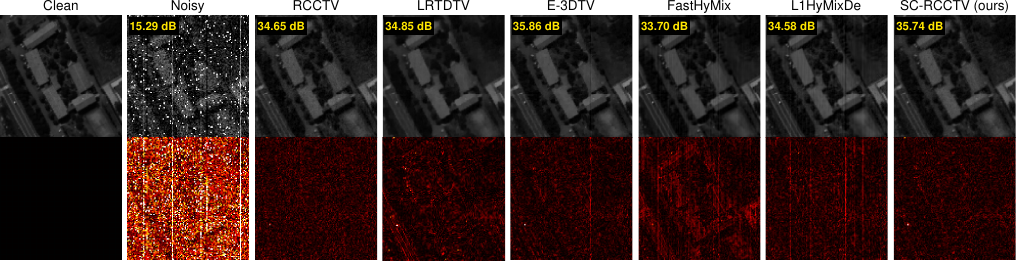}
\caption{Restoration results on Pavia University, case 8 (seed 0), band 50. Top: restored bands with the PSNR of the band; bottom: absolute error maps (display range $[0,0.25]$).}
\label{fig:sim}
\end{figure*}

\emph{Visual comparison:} Fig.~\ref{fig:sim} shows band 50 of Pavia University in case 8, which contains Gaussian noise, impulses and stripes. The error map of LRTDTV concentrates along edges, and those of FastHyMix and L1HyMixDe exhibit vertical lines at the striped columns. E-3DTV and SC-RCCTV yield the smallest errors in this band; SC-RCCTV leaves no visible residual lines, and its error map is close to spatially uniform.

\subsection{Real Data Experiments}\label{sec:real}
\begin{figure*}[!t]
\centering
\includegraphics[width=\textwidth]{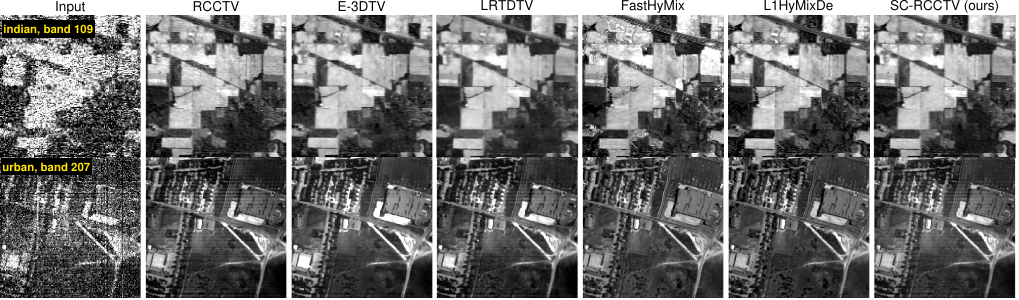}
\caption{Restoration results on real data. Top: Indian Pines band \nm{pick.indian}, the noisiest band outside the water-absorption windows; bottom: Urban band \nm{pick.urban}, a heavily striped band. Display range common to each row: 1st--99th percentile of all restored versions of that band.}
\label{fig:real}
\end{figure*}

Two real HSIs are used: Indian Pines (AVIRIS, $145{\times}145{\times}220$, water-absorption bands kept) and Urban (HYDICE, 210 bands; a $200{\times}200$ crop and the full $307{\times}307$ scene). Each band is min--max normalized. Since no reference is available, the Indian Pines results are evaluated by classification accuracy and the Urban results by a stripe index and the change of the band means. For classification, an RBF-SVM (one-versus-one, standardized features, automatic kernel scale) is trained on all bands with 10\% of the labeled pixels of each class (at least five) and tested on the remaining labeled pixels, over five random splits. The stripe index of a band is the mean absolute deviation of its column-mean profile from a seven-point moving median, plus the same quantity for the row-mean profile; on the $200{\times}200$ crop, it is averaged over the 20 bands of the input with the highest noise level (vertical-difference MAD). The largest change of a band mean, $\max_k|\operatorname{mean}(\hat X_k-Y_k)|$, indicates radiometric bias only; it does not measure the noise removed. Table~\ref{tab:real} and Fig.~\ref{fig:real} report the results.

\begin{table}[!t]
\caption{Real data. Indian Pines: overall accuracy (OA, mean $\pm$ std over 5 splits) and $\kappa$ of an RBF-SVM with 10\% training samples per class. Urban ($200{\times}200$): stripe index averaged over the 20 noisiest bands of the input, and largest change of a band mean, both $\times10^{-3}$. Time: Indian Pines, idle machine; t-CTV was run on Indian Pines only.}
\label{tab:real}
\centering
\scriptsize
\setlength{\tabcolsep}{3pt}
\begin{tabular}{lccccc}
\toprule
Method & OA (\%)$\uparrow$ & $\kappa\uparrow$ & Stripe$\downarrow$ & Shift$\downarrow$ & Time (s)\\
\midrule
Noisy input & 72.3\,$\pm$\,0.4 & 0.676 & 107.8 & 0 & --\\
RCCTV & 74.9\,$\pm$\,0.7 & 0.709 & 89.5 & 30.8 & 2.0\\
LRTDTV & 82.3\,$\pm$\,0.3 & 0.796 & 38.3 & 39.4 & 46\\
E-3DTV & 82.1\,$\pm$\,0.3 & 0.793 & 20.7 & 114.7 & 13\\
t-CTV & 89.2\,$\pm$\,0.5 & 0.875 & -- & -- & 270\\
FastHyMix & 80.1\,$\pm$\,0.3 & 0.771 & 12.2 & 20.4 & 3.7\\
L1HyMixDe & 79.0\,$\pm$\,0.6 & 0.758 & 11.0 & 37.3 & 12\\
SC-RCCTV & 78.1\,$\pm$\,0.5 & 0.746 & 49.7 & 9.4 & 4.6\\
\bottomrule
\end{tabular}
\end{table}

On Indian Pines, the overall accuracy (OA) of an RBF-SVM rises from \nm{oa.Input}\% (noisy input) and \nm{oa.RCCTV}\% (RCCTV) to \nm{oa.SC}\% (SC-RCCTV), and falls to \nm{oa.SC_noG}\% without the dense term and \nm{oa.SC_scalar}\% with a scalar $\sigma$, confirming that the band-wise noise level is the main source of the improvement. The remaining methods achieve higher accuracy: L1HyMixDe \nm{oa.L1HyMixDe}\%, FastHyMix \nm{oa.FastHyMix}\%, E-3DTV \nm{oa.E3DTV}\%, LRTDTV \nm{oa.LRTDTV}\% and t-CTV \nm{oa.tCTVd}\%. Increasing the rank from 8 to 16 lifts SC-RCCTV to \nm{rank.oa16}\% (\nm{rank.oa4}\% at rank 4, \nm{rank.oa32}\% at rank 32), which narrows but does not close the gap to FastHyMix. Visually (Fig.~\ref{fig:real}, top), SC-RCCTV recovers the field boundaries of band \nm{pick.indian} without visible artifacts, whereas FastHyMix leaves bright speckles in the upper part of the scene.

On Urban, the stripe index decreases from \nm{si.Input} to \nm{si.SC} ($\times10^{-3}$; \nm{si.SC_noT} without the stripe term), but it remains higher than those of L1HyMixDe (\nm{si.L1HyMixDe}), FastHyMix (\nm{si.FastHyMix}) and E-3DTV (\nm{si.E3DTV}). Faint residual horizontal lines remain in the output of SC-RCCTV (Fig.~\ref{fig:real}, bottom): the Urban stripes do not extend along entire rows, and by \eqref{eq:fatest}, a stripe that covers only a fraction $f$ of a row contributes at most $f$ times its amplitude to the mean score, so that it may remain below the detection threshold. On the other hand, SC-RCCTV changes the band means least (largest shift \nm{sh.SC}, against \nm{sh.FastHyMix} for FastHyMix and \nm{sh.E3DTV} for E-3DTV, $\times10^{-3}$). In the near-dead water-absorption bands (e.g., Indian Pines band 151 and Urban band 147), FastHyMix and L1HyMixDe return sharp images synthesized from their spectral subspace, whereas the low-rank and TV models, SC-RCCTV included, return smooth images; without a reference, neither result can be verified.

\subsection{Discussion}\label{sec:disc}

\subsubsection{Effectiveness of Each Component}\label{sec:abl}
Table~\ref{tab:abl} analyzes the contribution of each term. Without the stripe term, SC-RCCTV loses \nm{u.SC_noT.st}\,dB on the stripe cases; without the dense term, it loses \nm{u.SC_noG.gs}\,dB on cases 1--3; with a scalar $\sigma$ instead of $\sigma_k$, it loses \nm{u.SC_scalar.hv}\,dB on the band-varying cases (\nm{w.SC_scalar.hv} cells). Modeling only column stripes gives the same MPSNR on these column-striped data (difference \nm{d.SCv.all}\,dB), so including the row component does not degrade the results.

\begin{table}[!t]
\caption{Effectiveness of each component: mean MPSNR gain (dB) over RCCTV in each noise family of Table~\ref{tab:cases} (both cubes, all realizations). Last row: the same model with the true $\sigma_k$ in place of the estimated ones.}
\label{tab:abl}
\centering
\scriptsize
\setlength{\tabcolsep}{4pt}
\begin{tabular}{lcccc}
\toprule
Variant & G+I & Str. & B.-v. & All\\
\midrule
SC-RCCTV (full) & +1.74 & +2.64 & +1.94 & +2.24\\
without stripe term $T$ & +1.75 & +0.08 & +1.06 & +0.77\\
without dense term $G$ & $-$0.01 & +1.14 & +0.36 & +0.65\\
scalar $\sigma$ instead of $\sigma_k$ & +1.76 & +2.63 & +0.43 & +1.98\\
$\hat\sigma_k^{(0)}$ kept fixed (two-stage) & +1.68 & +2.18 & +1.80 & +1.96\\
\midrule
true $\sigma_k$ (all else unchanged) & +1.76 & +2.65 & +1.94 & +2.25\\
\bottomrule
\end{tabular}
\end{table}

Two further comparisons concern the estimation of the noise levels. First, keeping the initial estimate $\hat\sigma_k^{(0)}$ fixed, as in a two-stage scheme that estimates the noise levels before denoising, lowers the MPSNR by \nm{u.SC_fix.all}\,dB on average (SC-RCCTV is better in \nm{w.SC_fix.all} cells and at most \nm{umin.SC_fix}\,dB worse in the others). Most of this gain (\nm{fix.share}\%) comes from case~4, whose bands contain no Gaussian noise, so that the image structure dominates the initial estimate (\nm{fix.c4}\,dB); in the other cases it is \nm{fix.other}\,dB. The MSSIM changes by at most \nm{ufix.mssim.max}, while the SAM is lower with the fixed estimate, by \nm{ufix.sam.belt100} on Beltsville and \nm{ufix.sam.pavia120} on Pavia University. On the real data, the fixed estimate lowers the OA on Indian Pines from \nm{oa.SC}\% to \nm{oa.SC_fix}\% and raises the stripe index on Urban from \nm{si.SC} to \nm{si.SC_fix} ($\times10^{-3}$). Second, the same model with the true $\sigma_k$ in place of the estimates, all else unchanged, differs from SC-RCCTV by \nm{u.SC_true.all}\,dB on average and by at most \nm{amax.SC_true}\,dB in any cell (last row of Table~\ref{tab:abl}), so that little is lost by estimating the noise levels.

\begin{figure}[!t]
\centering
\includegraphics[width=\columnwidth]{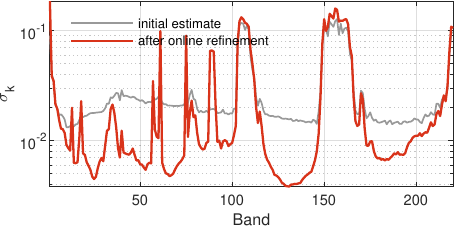}
\caption{Noise levels $\hat\sigma_k$ estimated by SC-RCCTV on Indian Pines: the stripe-invariant initial estimate $\hat\sigma_k^{(0)}$ and the value after the online refinement (logarithmic scale).}
\label{fig:sigma}
\end{figure}

Fig.~\ref{fig:sigma} illustrates the online refinement on Indian Pines. Outside the water-absorption windows, the refined noise level is lower than the initial estimate in most bands (median ratio \nm{sig.medratio}, up to \nm{sig.maxratio}), because edges and texture inflate the MAD of the differences; it is more than twice the initial value only in bands \nm{sig.bands2x}, where the rank-8 representation presumably leaves part of the signal in the residual. In the water-absorption bands, where the noise dominates, the two estimates agree within \nm{sig.absdev}\%.

On stripe-free Gaussian cubes (three $64{\times}64$ cubes, $\sigma\in\{0.05,0.1\}$), the fraction of nonzero stripe amplitudes is \nm{fa.c3}\% for $c{=}3$ and \nm{fa.c2}\% for $c{=}2$. For these cubes, $\eta_k\approx1.6$ and $\gamma(\eta_k)\approx0.83$, and \eqref{eq:farate} predicts \nm{fapred.c3}\% and \nm{fapred.c2}\%, in agreement with a Monte Carlo simulation of the exact column problem ($0.095$--$0.099\%$ and $2.78$--$2.88\%$); the deviations of the measured rates can therefore be attributed to the estimated noise levels and the finite number of iterations. The estimated noise levels have nearly settled at termination: over \nm{fp.runs} runs on $64{\times}64$ crops, the median of the largest relative change of any $\hat\sigma_k$ in the last iteration is $\nm{fp.med}$ (worst case $\nm{fp.max}$), and SC-RCCTV terminates after \nm{iters.SC} iterations on average on the native cubes. Continuing to a tolerance of $10^{-7}$ (\nm{iters.SC_tight} iterations on average) changes the MPSNR by \nm{tight.mean}\,dB on average and by at most \nm{tight.max}\,dB over \nm{tight.n} cells.

\subsubsection{Computational Efficiency}
\begin{figure*}[!t]
\centering
\includegraphics[width=\textwidth]{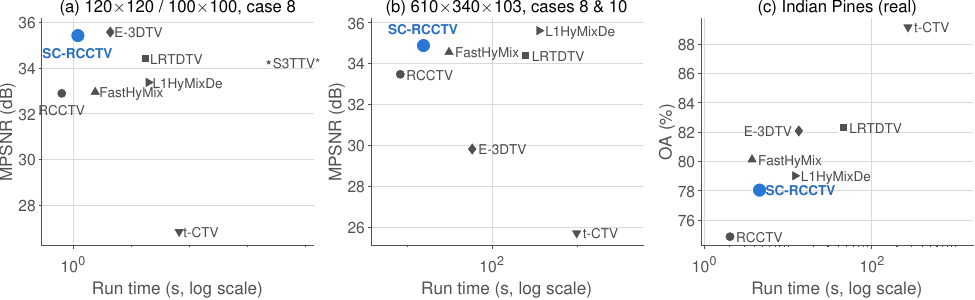}
\caption{Accuracy against run time (log scale). (a) Native-size cubes, case 8, mean of both cubes; (b) full Pavia University, mean of cases 8 and 10; (c) Indian Pines, SVM overall accuracy. S3TTV: run time from the per-iteration time.}
\label{fig:pareto}
\end{figure*}

On the native cubes (Table~\ref{tab:avg}), SC-RCCTV takes \nm{t.SC.pavia120_c8}\,s and \nm{t.SC.belt100_c8}\,s, against \nm{t.RCCTV.pavia120_c8}\,s and \nm{t.RCCTV.belt100_c8}\,s for RCCTV and \nm{t.FastHyMix.pavia120_c8}\,s and \nm{t.FastHyMix.belt100_c8}\,s for FastHyMix, while E-3DTV, LRTDTV, t-CTV and S3TTV require \nm{trr.E3DTV}, \nm{trr.LRTDTV}, \nm{trr.tCTVd} and \nm{trr.S3TTV} times as long as SC-RCCTV, respectively. To verify that this ranking holds at scene size, all methods except S3TTV (whose memory exceeds 16\,GB) were also run on the full Pavia University scene ($610{\times}340{\times}103$; cases 8 and 10) and on the full Urban scene ($307{\times}307{\times}210$). On Pavia University, SC-RCCTV reaches \nm{L.SC.c8} and \nm{L.SC.c10}\,dB in \nm{L.SC.tpavia}\,s; FastHyMix reaches \nm{L.FastHyMix.c8} and \nm{L.FastHyMix.c10}\,dB in \nm{L.FastHyMix.tpavia}\,s, and L1HyMixDe \nm{L.L1HyMixDe.c8} and \nm{L.L1HyMixDe.c10}\,dB (the best value on case 10) in \nm{L.L1HyMixDe.tpavia}\,s; the results of the other methods are shown in Fig.~\ref{fig:pareto}(b). On the full Urban scene, SC-RCCTV lowers the stripe index of RCCTV from \nm{L.RCCTV.urbanst} to \nm{L.SC.urbanst} ($\times10^{-3}$) in \nm{L.SC.turban}\,s, but the other methods reach lower values (e.g., \nm{L.FastHyMix.urbanst} for FastHyMix and \nm{L.E3DTV.urbanst} for E-3DTV), as on the $200{\times}200$ crop. Fig.~\ref{fig:pareto} summarizes the trade-off between accuracy and run time.

\subsubsection{Applicability and Limitations}
SC-RCCTV is most beneficial when stripes extend along entire columns or rows, as produced by detector offsets in push-broom sensors: it removes them without prior knowledge of their number or amplitude and, on the native-size cubes, outperforms the parameter-free subspace methods by several decibels, in about half the run time of FastHyMix (also on the full Pavia University scene, \nm{L.SC.tpavia} vs.\ \nm{L.FastHyMix.tpavia}\,s). On the full Pavia University scene, the margin in case 8 is small, and in case 10 L1HyMixDe is the most accurate. In the absence of stripes, FastHyMix and L1HyMixDe retain an advantage, most clearly for Gaussian noise alone, and on the two real scenes, whose stripes are only partly line-constant, FastHyMix attains higher classification accuracy and stronger destriping. The two classes of methods are therefore complementary. Several aspects of the model deserve further study: the rank-8 spatial representation limits the recovery of near-dead bands; the line-constant stripe model does not capture stripes that cover only part of a line, and it cannot always distinguish stripes from image structures that are themselves nearly constant along a line, such as straight roads or field boundaries; and the convergence of the iteration with adaptive noise levels, although consistently observed, remains to be proven.

%=====================================================================================================
\section{Conclusion}\label{sec:concl}
This article proposed SC-RCCTV, a self-calibrated RCCTV model for HSI mixed noise removal, in which a band-wise dense term and line-constant stripe terms are introduced and all noise parameters are estimated within the model: the noise levels by a stripe-invariant MAD estimate refined online from the residual, and the stripe thresholds by a significance test with an analytically predictable false-alarm rate. The resulting ADMM-based algorithm has closed-form subproblems and the same per-iteration complexity as RCCTV. Experiments show that SC-RCCTV stays close to the same model given the true noise levels, that estimating the noise levels within the model improves on estimating them beforehand, and that under line-constant stripes it outperforms FastHyMix and L1HyMixDe by several decibels on native-size cubes and S3TTV, which is given the true noise parameters, by \nm{s0u.S3TTV.st}\,dB, at a run time of about one second. Future work will extend the stripe model to segment-wise offsets, detected by the same significance test, so as to handle real stripes that cover only parts of lines. The MATLAB code, the noisy inputs and all result tables are available at \codeurl.

\section*{Acknowledgment}
The author thanks the authors of RCCTV, LRTDTV, E-3DTV, t-CTV, FastHyMix, L1HyMixDe and S3TTV for releasing their codes, and the providers of the Pavia University, Beltsville, Indian Pines, Urban and Washington DC Mall data.

\bibliographystyle{IEEEtran}
\bibliography{refs}

\end{document}